\documentclass[conference]{IEEEtran}
\IEEEoverridecommandlockouts
\usepackage{cite}
\usepackage{amsmath,amssymb,amsfonts}
\usepackage{algorithmic}
\usepackage{graphicx}
\usepackage{textcomp}
\usepackage{xcolor}
\usepackage{hyperref}
\usepackage{microtype}
\usepackage{subfigure}
\usepackage{booktabs} 
\usepackage{wrapfig}
\usepackage{makecell}

\def\BibTeX{{\rm B\kern-.05em{\sc i\kern-.025em b}\kern-.08em
    T\kern-.1667em\lower.7ex \hbox{E}\kern-.125emX}}
\begin{document}

\title{Adaptive Multi-receptive Field Spatial-Temporal Graph Convolutional Network for Traffic Forecasting
}

\author{

\IEEEauthorblockN{Xing Wang$^1$, Juan Zhao$^1$, Lin Zhu$^1$, Xu Zhou$^2$, Zhao Li$^2$,
Junlan Feng$^1$, Chao Deng$^1$, Yong Zhang$^2$}
\IEEEauthorblockA{$^1$ China Mobile Research Institute, Beijing, China}
\IEEEauthorblockA{$^2$ Electronic Engineering, Beijing University of Posts and Telecommunications, Beijing, China}
\IEEEauthorblockA{
$^1$\{wangxing, zhaojuan, zhulinyj, fengjunlan, dengchao\}@chinamobile.com,
$^2$\{zhou\_xu, L\_zhao, yongzhang\}@bupt.edu.cn}


}

\maketitle

\begin{abstract}
Mobile network traffic forecasting is one of the key functions in daily network operation. A commercial mobile network is large, heterogeneous, complex and dynamic. These intrinsic features make mobile network traffic forecasting far from being solved even with recent advanced algorithms such as graph convolutional network-based prediction approaches and various attention mechanisms, which have been proved successful in vehicle traffic forecasting. In this paper, we cast the problem as a spatial-temporal sequence prediction task. We propose a novel deep learning network architecture, Adaptive Multi-receptive Field Spatial-Temporal Graph Convolutional Networks (AMF-STGCN), to model the traffic dynamics of mobile base stations. AMF-STGCN extends GCN by (1) jointly modeling the complex spatial-temporal dependencies in mobile networks, (2) applying attention mechanisms to capture various Receptive Fields of heterogeneous base stations, and (3) introducing an extra decoder based on a fully connected deep network to conquer the error propagation challenge with multi-step forecasting. Experiments on four real-world datasets from two different domains consistently show AMF-STGCN outperforms the state-of-the-art methods.
\end{abstract}

\begin{IEEEkeywords}
traffic forecasting, spatial-temporal data, Graph Convolutional Network, mobile traffic, deep learning
\end{IEEEkeywords}

\section{Introduction}
\label{Introduction}

Total global mobile data traffic reached 58EB per month at the end of 2020. It is projected to grow monthly by 5 percent in coming years. Data traffic prediction is becoming of fundamental importance for 4G/5G telecom network operation. Accurate traffic forecast enables key parameters of base stations to be automatically adapted for optimal user experience and energy consumption. With large-scale commercial deployment of 5G network, traffic prediction has become one of the key enabling technologies for autonomous network, which is promoted by the entire telecom industry. In addition, traffic forecasting provides important support for many transportation services, such as traffic control, route planning, and navigation, etc \cite{yu2017spatio}.

\begin{figure}[htpb] 
\centering
\subfigure[Base stations in the wireless network.]{
\label{fig1-1}
\includegraphics[width=8cm]{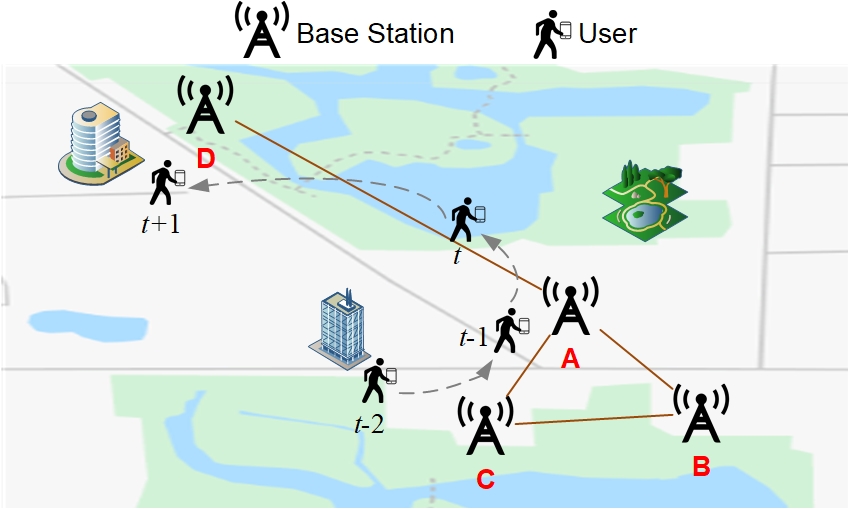}}
\subfigure[Complex spatial and temporal correlations in traffic network.]{
\label{fig1-2}
\includegraphics[width=8cm]{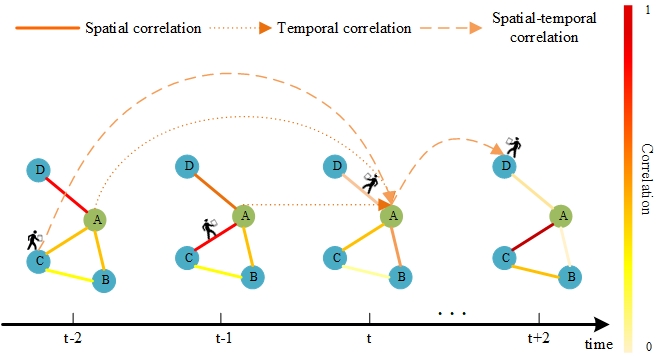}}
\caption{Spatial-temporal correlations in the mobile traffic network.}
\label{fig1}
\end{figure}

A typical telecom operator operates millions of base stations. Each base station connects with users within a certain radius range and transfers data for them. The generated data traffic shares similar patterns with the road traffic data. First, traffic data of mobile network or road network is inherent with complex spatial-temporal dependencies. Fig.~\ref{fig1-1} shows the user migration between base stations, resulting in explicit or implicit spatial-temporal correlations among the base stations. Due to the migration of users, the data traffic of a base station $A$ at time step $t$, represented as $A_t$, is influenced by $B_t$, $C_t$ and $D_t$ in the spatial domain, and the impacts change over time as the bold lines shown in Fig.~\ref{fig1-2}. Meanwhile, $A_t$ correlates with its own past traffic $A_{0,....,t-1}$ in the time axis.
In addition, the information can propagate along the spatial and temporal dimensions simultaneously. And the propagation process can be discontinuous due to the uncertainty of wireless signal, user behavior, environment change, etc. For example, as shown in Fig.~\ref{fig1-2}, a user can consume a large volume of data via a given base station $C$ at time step ${t} - 2$. At the next moment, this user can stop the connection or migrates to a different base station $A$ without consuming any data traffic until time step $t$. Then the traffic volume of $C_{t-2}$ actually affects the traffic volume of $A_t$ directly. The discontinuous nature of spatial-temporal data generation results in the inherent non-local spatial-temporal synchronization dependencies along with the current and history moments. And we believe that if we capture the complex spatial-temporal synchronization correlations directly, the traffic prediction performance can be improved significantly.

In this paper, we frame the problem as a Spatial-Temporal series prediction task, where both input and output are spatial-temporal data series. Base stations form a graph in space based on distances between them and business reliances, where each node corresponds to a base station and each edge represents their relationship.

Traditional time series modeling algorithms such as the autoregressive integrated moving average (ARIMA) \cite{williams2003modeling} and its variants are hard to be extended to represent spatial correlations. Researchers have proposed various deep learning models to capture the complex reliances hidden in traffic data of the transportation domain, which has similarity to mobile traffic, though mobile traffic data are relatively more dynamic and unstable. Most of these models use a few building blocks to respectively capture spatial dependencies, temporal dependencies and then fuse embeddings from both spatial and temporal domains. Graph convolutional networks (GCN) is one of the most consolidated approaches to capture the spatial correlations among nodes,  while CNN,  recurrent neural networks including Gate Recurrent Units (GRU) and long-short term memory (LSTM) have been used to model temporal dynamics \cite{yu2017spatio} \cite{li2017diffusion}. \cite{guo2019attention} 
introduced attention mechanisms including spatial attention and temporal attention to enhance GCN and GNN to better model the nonlinearities. These models have advanced the technology and shown significant improvement on prediction performance.

However, two remaining challenges are not addressed fairly. One, most aforementioned models capture spatial correlations and temporal correlations separately, and then fuse them together. We argue the complex spatial-temporal correlations should be synchronously modeled. STSGCN \cite{song2020spatial} used multiple local spatial-temporal graphs to model the spatial-temporal synchronous correlations of the local adjacent time steps. But STSGCN only focused on the localized spatial-temporal correlations without considering the discontinuous nature of spatial-temporal data generation. Hence, in this paper we propose a novel spatial-temporal joint graph convolutional network to directly capture the intrinsic non-local spatial-temporal reliances.

\begin{figure}[htpb] 
\centering
\subfigure[Mobile traffic data.]{
\label{fig2-1}
\includegraphics[width=4cm]{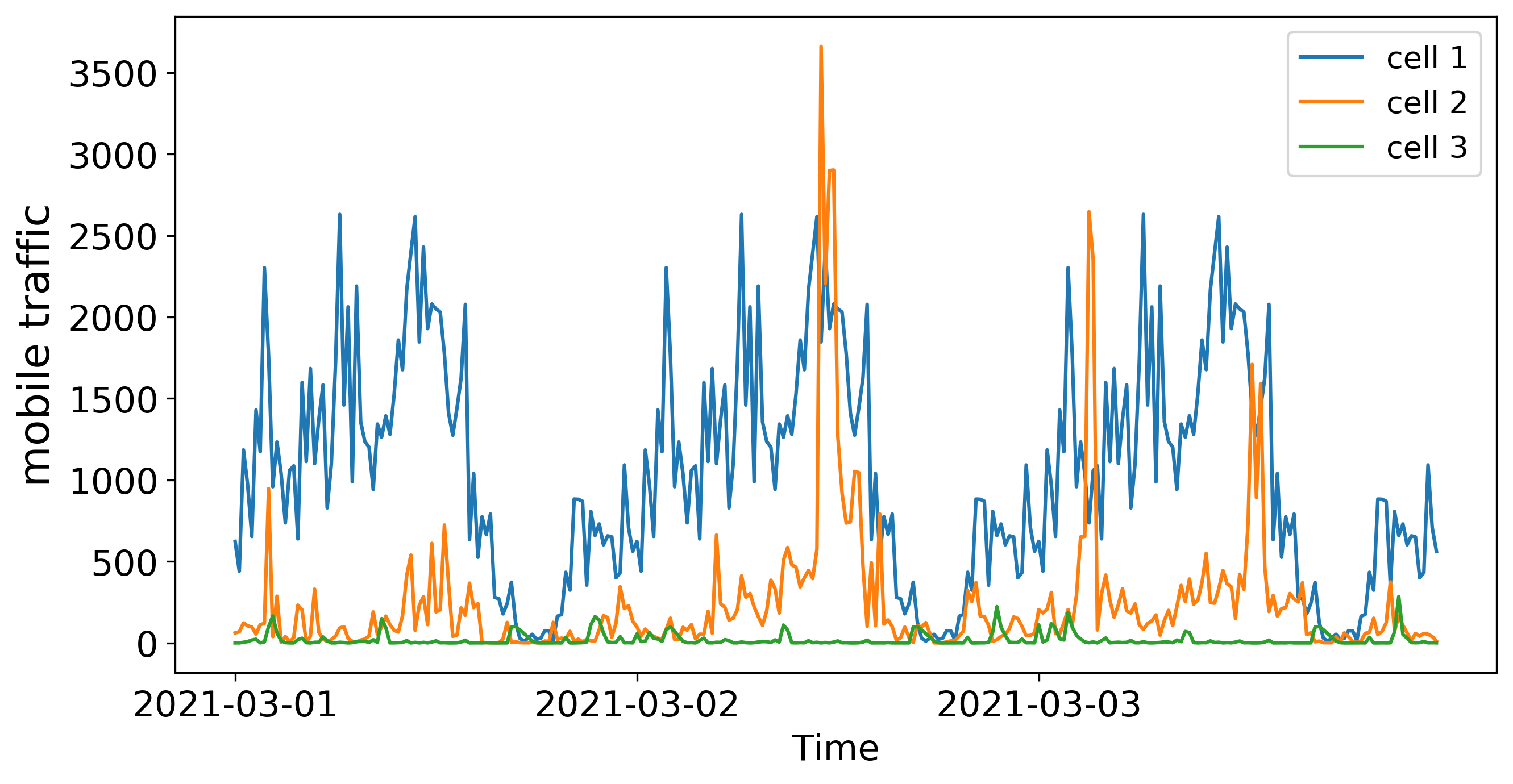}}
\subfigure[Road traffic data.]{
\label{fig2-2}
\includegraphics[width=4cm]{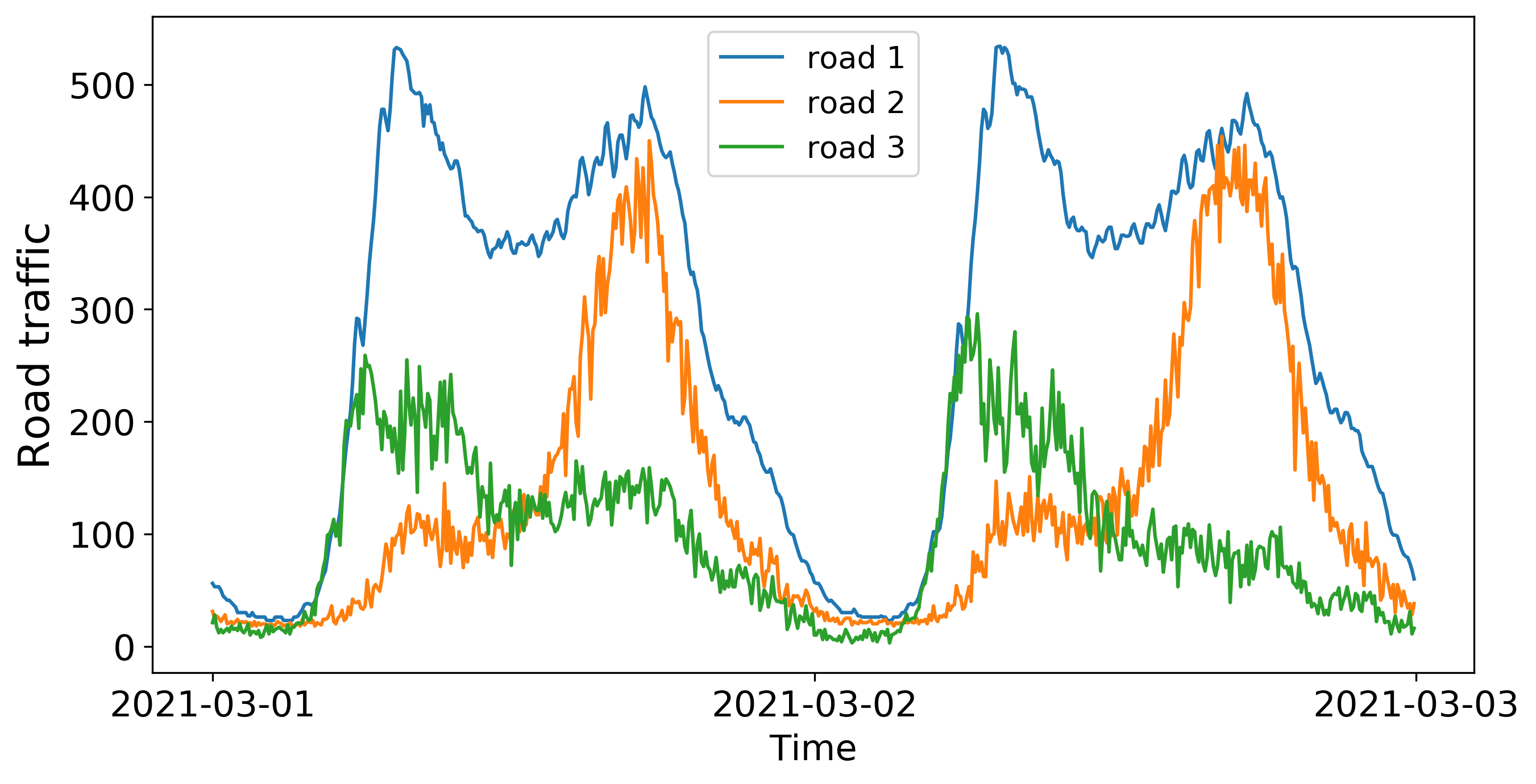}}
\caption{The heterogeneity of traffic data }
\label{fig2}
\end{figure}

Second, heterogeneity of nodes is not well considered. Especially in 5G, a base station could be a wide-area station, a medium range station, or a local area station. Their reliances with nearby stations are quite different. Heterogeneity is also an evident concern to model traffic network. As shown in Fig.~\ref{fig2}, the traffic data sequence of different nodes can behave quite variously in both the mobile network and road network. The traffic in an urban area is denser, which means the roads of such regions can be affected by a wider range of neighborhoods. On the contrary, the traffic of suburban area is relatively sparse. To address this challenge, we propose an Adaptive Multi-receptive Field Spatial-Temporal Graph Convolutional Block (AMF-STConv Block), which contains the multi-scale spatial-temporal feature extraction and a graph attention module to adaptively model the graph. Following the multiple AMF-STConv Blocks, we design a fusion module with two decoders named Fully-Connected decoder and Time-Step decoder along with the pre-training mechanism to establish the relationships between the historical and future time steps to alleviate multi-step error accumulation.

In this paper, we propose a novel Adaptive Multi-receptive Field Spatial-Temporal Graph Convolutional Networks (AMF-STGCN) to address the above mentioned issues of traffic forecasting. 
To the best of our knowledge, we are the first to apply GCN to mobile network traffic prediction. 
Our main technical contributions are summarized as follows:

\begin{itemize}
\item We propose a novel deep learning network architecture, Adaptive Multi-receptive Field Spatial-Temporal Graph Convolutional Networks (AMF- STGCN) to model mobile network traffic of base stations.

\item Our proposed model, AMF-STGCN, is able to effectively model the complex spatial-temporal dependencies in mobile networks, capture the heterogeneity of base stations adaptively with an attention mechanism, and tackle the error propagation challenge with multi-step forecasting.

\item We evaluate AMF-STGCN on four real-world datasets from two fields and the experimental results prove AMF-STGCN achieves the best overall prediction performance comparing to current state-of-the-art prediction models.

\end{itemize}

\section{Methodology}

\subsection{Preliminaries}

In this paper, we define a static undirected  graph $\mathcal{G}=\left( V,E,A \right)$ as the traffic network. $V$ represents the set of nodes, $\left| V \right|=N$($N$ indicates the number of nodes). $E$ is the set of edges representing the connectivity between nodes. $A\in {{\mathbb{R}}^{N\times N}}$ is the adjacency matrix of $\mathcal{G}$, where ${{A}_{{{v}_{i}},{{v}_{j}}}} \in \{0, 1\}$ represents the connection between nodes ${{v}_{i}}$ and ${{v}_{j}}$. The graph signal matrix is expressed as ${{X}_{t}}\in {{\mathbb{R}}^{N\times C}}$, where $t$ denotes the timestep and $C$ indicates the feature dimension. The graph signal matrix represents the observations of graph network $\mathcal{G}$  at time step $t$.

\textbf{\textbf{Problem Defined}}   

Given the graph signal matrix of historical $T$ time steps $X =\left( {{X}_{{{t}_{1}}}},{{X}_{{{t}_{2}}}},\ldots ,{{X}_{{{t}_{T}}}} \right)\in {{\mathbb{R}}^{T\times N\times C}}$, our goal is to predict the graph signal matrix of the next $M$ time steps $\hat{Y}=\left( {{{\hat{X}}}_{{{t}_{T+1}}}},{{{\hat{X}}}_{{{t}_{T+2}}}},...,{{{\hat{X}}}_{{{t}_{T+M}}}} \right)\in {{\mathbb{R}}^{M\times N\times C}}$. Given an undirected graph $\mathcal{G}$, we need to learn a mapping function $\mathcal{F}$ to map the graph signal matrix of historical time steps to the future time steps: 
\begin{equation}
    \left(\hat{X}_{t_{T+1}}, \hat{X}_{t_{T+2}}, \cdots, \hat{X}_{t_{T+M}}\right)=\mathcal{F}_{\theta}\left(X_{t_{1}}, X_{t_{2}}, \cdots, X_{t_{T}};\mathcal{G}\right)
\end{equation}

\noindent where $\theta$ represents learnable parameters of our model.

\subsection{Architecture}

\begin{figure}[htpb] 
\centering
\subfigure[The proposed AMF-STGCN, which consists of AMF-STConv Blocks and fusion module.]{
\label{fig3-1}
\includegraphics[width=6cm]{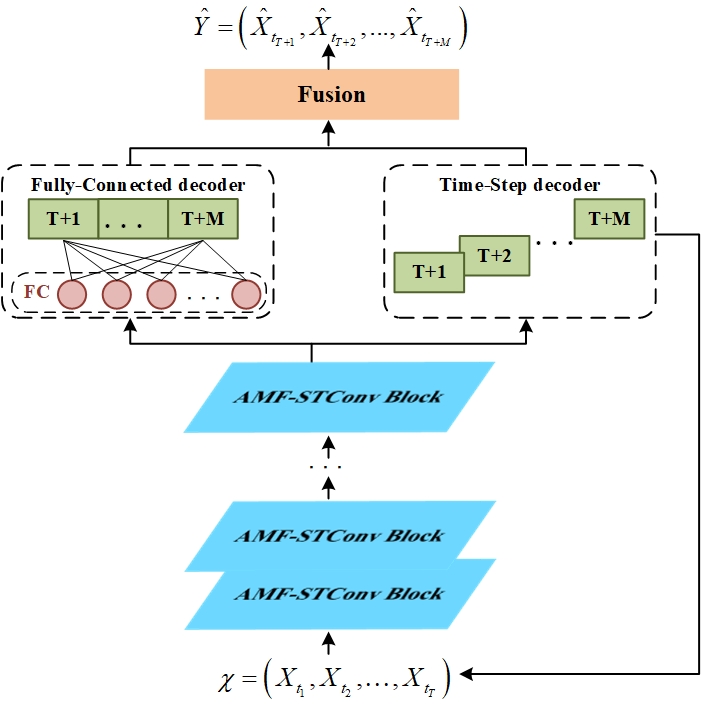}}
\subfigure[The architecture of AMF-STConv Block.]{
\label{fig3-2}
\includegraphics[width=5cm]{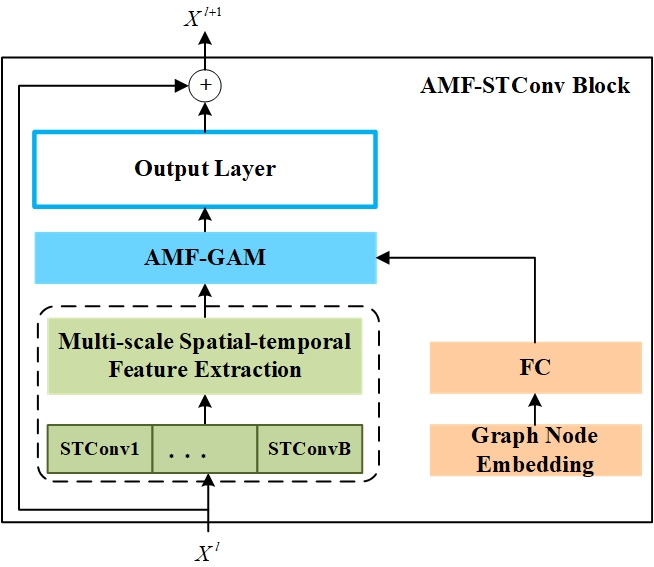}}
\caption{Overall architecture of AMF-STGCN.}
\label{fig3}
\end{figure}

The architecture of the AMF-STGCN is shown in Fig.~\ref{fig3-1}, which consists of multiple Adaptive Multi-receptive Field Spatial-Temporal Graph Convolutional Blocks (AMF-STConv block), a Fully-Connected decoder, a Time-Step decoder, and a Fusion module. Each AMF-STConv block is composed of the proposed spatial-temporal joint graph convolution with adaptive receptive field graph attention mechanism, to adaptively model the graph heterogeneity. We also incorporate two decoders and the pre-training mechanism into the model to achieve accurate multi-step traffic forecast.

\subsection{Adaptive Multi-Receptive Field Spatial-Temporal Graph Convolutional Block}

The AMF-STConv block, as shown in Fig.~\ref{fig3-2}, first implements multiple spatial-temporal joint convolution kernels, such as $B$ branches (kernels), to achieve multi-receptive field spatial-temporal features extraction. Then the extracted features go through an Adaptive Multi-receptive Field Graph attention module (AMF-GAM), which combine graph node embedding with the attention to achieve the heterogeneity modeling. At last, we use a two-layer fully connected network (the output layer) to produce the output. We denote the input of the ${l^{th}}$ AMF-STConv block as ${X^l}$, and the output as ${X^{l + 1}}$.

\textbf{Spatial-temporal joint graph convolution}

Previous models captured spatial and temporal features with separate modules respectively, which ignored the synchronous correlation \cite{yu2017spatio, AGCRN, li2017diffusion, guo2019attention, he2019graph}. In order to extract spatial-temporal correlations simultaneously, we propose spatial-temporal joint Graph Convolution (STConv). In this paper, we construct STConv based on graph convolution in the spectral domain, which is implemented by using the graph Fourier transform basis from eigenvalue decomposition of the Laplacian matrix. Then it transforms the graph signals from spatial into the spectral domain.
To reduce the computation complexity, Chebyshev polynomial ${T_k}\left( x \right)$ is used for approximation. The spectral graph convolution can be written as \cite{kipf2016semi, yu2017spatio}:

\begin{equation}
    \Theta { * _{\cal G}}x = \Theta \left( L \right)x \approx \sum\limits_{k = 0}^{K - 1} {{\theta _k}{T_k}\left( {\tilde L} \right)x}
\end{equation}

\noindent where ${{*}_{\mathcal{G}}}$ is graph convolutional operator, $\Theta $ is graph convolution kernel, $x\ \in \ {{\mathbb{R}}^{N}}$ is the graph signal, ${{T}_{k}}\left( {\tilde{L}} \right)\ \in \ {{\mathbb{R}}^{N\times N}}$ is the Chebyshev polynomial of order \textit{k} with the scaled Laplacian $\tilde{L}=\frac{2}{{{\lambda }_{max}}}L-{{I}_{N}}$ ($L$ is the normalized graph  Laplacian, ${{\lambda }_{max}}$ is the largest eigenvalue of $L$, ${{I}_{N}}$ is identity matrix). ${{\theta }_{k}}$ is the coefficient of the \textit{k}-th order polynomial.

\begin{figure}[htpb] 
\centering
\subfigure[Two STConv kernel examples.]{
\label{fig4-1}
\includegraphics[width=4.5cm]{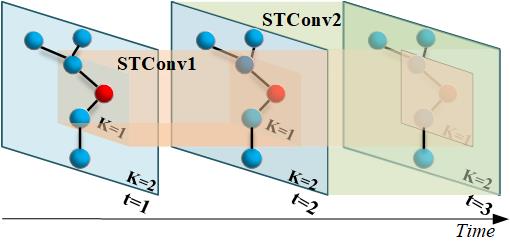}}
\subfigure[A schematic diagram of STConv kernel.]{
\label{fig4-2}
\includegraphics[width=3.5cm]{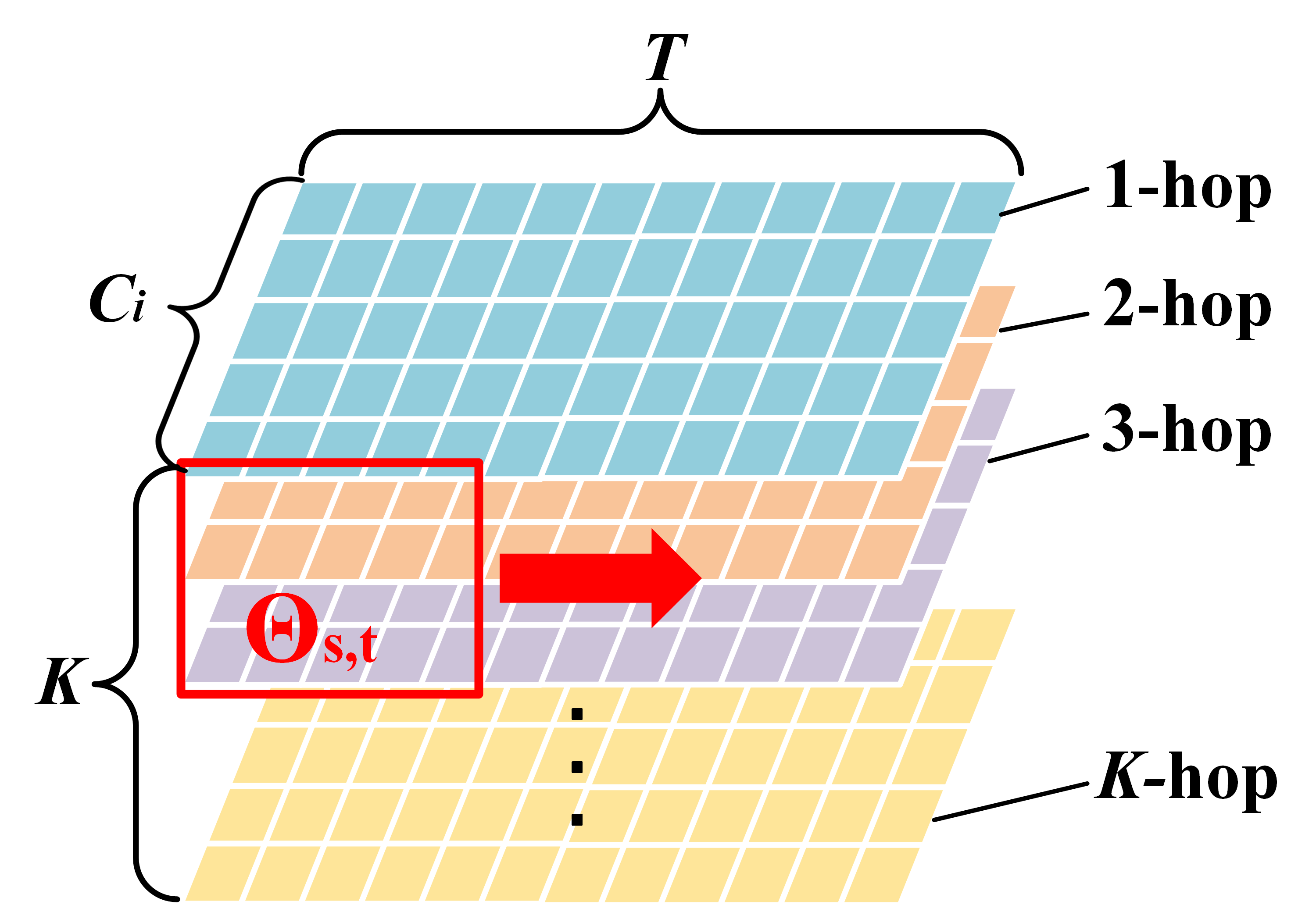}}
\caption{Spatial-temporal joint graph convolution.}
\label{fig4}
\end{figure}

Based on spectral domain graph convolution, as shown in Fig.~\ref{fig4} (b), we concatenate the \textit{K}-hop ${{T}_{k}}\left( {\tilde{L}} \right)$ to form the graph convolution result with multi-scale spatial receptive fields. ${K}$ indicates the furthest receptive field in the spatial dimension. Different from \cite{3DGCN} which only calculated the ${{T}_{K}}\left( {\tilde{L}} \right)$ graph convolution without the concatenation, all nodes in the graph share the same $K$th-hop only spatial receptive field without multi-scale. Then we construct the spatial-temporal joint graph convolution kernel ${\Theta _{s,t}} \in {{\mathbb{R}}^{{K_t} \times {K_s} \times {C_i} \times {C_o}}}$, where ${K_t}$ and ${{K}_{s}}$ represents the kernel size in the temporal and spatial
dimension respectively, $C_i$ is the number of input channels, $C_o$ is the number of filters. And the spatial-temporal joint graph convolution can be formulated as:

\begin{equation}
    {{\bf{T}}_{\bf{K}}}\left( {\tilde{L}} \right) = Concat({T_0}\left( {\tilde{L}} \right),{T_1}\left( {\tilde{L}} \right), \ldots ,{T_{K - 1}}\left( {\tilde{L}} \right))
\end{equation}
\begin{equation}
    {\Theta _{s,t}} * {X^l} = {\Theta _{s,t}} * {{\bf{T}}_{\bf{K}}}\left( {\tilde L} \right)\;{X^l}
\end{equation}

\noindent where ${{\mathbf{T}}_{\mathbf{K}}}\left( {\tilde{L}} \right)\ \in \ {{\mathbb{R}}^{K\times N\times N}}$ is the concatenation of all Chebyshev polynomials in ($K$-1) hop, $*$ is the convolution operation between ${{\Theta }_{s,t}}$ and ${X^l}$, ${X^l} \in {{\mathbb{R}}^{N \times T \times {C_{\text{i}}}}}$ is the input of the ${l_{th}}$ AMF-STConv block, ${T}$ is the input time steps. The concatenated graph convolution result can be ${{\bf{T}}_{\bf{K}}}\left( {\tilde L} \right)\;{X^l} \in {\mathbb{R}^{N \times T \times K \times {C_i}}}$. After the spatial-temporal joint graph convolution without padding, the output can be written as ${F} \in {{\mathbb{R}}^{N \times (T - {K_t} + 1) \times (K - {K_s} + 1) \times {C_o}}}$. 

So the STConv kernel ${{\Theta }_{s,t}}$ has the local spatial-temporal receptive field of ${{K}_{t}}\times {{K}_{s}}$, and ${{K}_{s}}$ should be smaller than $K$ because of the defined largest graph convolution of $K$-hop. For example, as the STConv1, STConv2 shown in Fig.~\ref{fig4-1}, STConv1 represents the STConv kernel with size of $3 \times 2$ (${K_t} \times {K_s}$), which can be represented as ${\theta _{s,t}} \in {{\mathbb{R}}^{3 \times 2 \times {C_{i}} \times {C_{o}}}}$. This means it can extract the spatial-temporal features of the node itself and its 2-hop spatial neighbors in the three adjacent time steps by one layer, which is needed by STSGCM with above two layers in STSGCN \cite{song2020spatial}. And deep Graph Convolutional Network (GCN) is prone to cause the over smoothing problem. 

Besides, since the spatial neighbors have various influences on the central node, so we implement a learnable spatial mask matrix ${{W}_{mask}}\in {{\mathbb{R}}^{N\times N}}$ \cite{song2020spatial} to adjust the graph adjacency relationship for assigning weights to different neighbors. Like \cite{song2020spatial}, we do the element-wise product between adjacency matrix $A$ and ${{W}_{mask}}$ to build a weight adjusted adjacency matrix: ${A' = }{{W}_{mask}} \otimes A \in {{\mathbb{R}}^{N \times N}}$. ${A'}$ is used to compute all graph convolutions in AMF-STGCN.

\begin{figure}[htpb]
\begin{center}
\includegraphics[width=7cm]{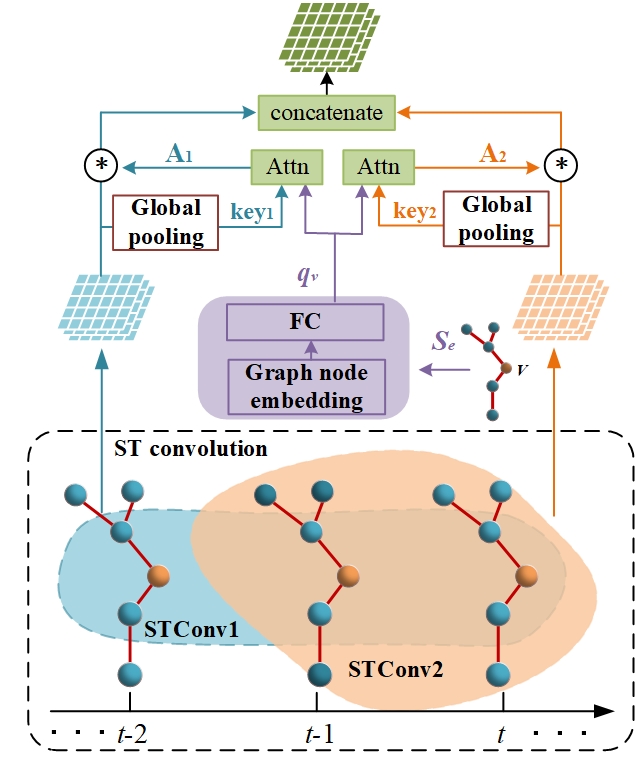}
\end{center}
\caption{Adaptive Multi-Receptive Field Graph Attention Module.}
\label{fig5}
\end{figure}

\textbf{Adaptive Multi-Receptive Field Graph Attention Module}

Different from the images whose data distribution is consistent, each node of the graph usually represents a road segment or a base station etc. Affected by external factors such as geographic location and surrounding environment, the traffic data of the graph nodes are various, namely the heterogeneity, as shown in Fig.~\ref{fig2}. To address this problem, an intuitive method is to learn the individual model for each node, but this method could cause extensive parameters and is hard to generalize. As a result, we aim to build a unified model to achieve traffic forecasting with heterogeneity modeling. Due to the nodes’ various properties and local spatial structures, we understand that a key manifestation of heterogeneity is the different local spatial-temporal receptive fields of each graph node. This means that different graph nodes have different perceptions of multiple spatial-temporal receptive fields. Inspired by \cite{zheng2020gman, AGCRN}, we apply a learnable graph node embedding ${S_e} \in {{\mathbb{R}}^{N \times d}}$ to represent the properties of each node in high-dimensional space, where $d$ represents the node embedding dimension. Meanwhile, we introduce inception \cite{szegedy2015going} to build multi-scale spatial-temporal receptive fields. Then, the graph node-level attention combined with multi-scale spatial-temporal receptive fields is proposed to adaptively model the graph heterogeneity.

The entire architecture of Adaptive Multi-receptive Field Graph Attention Module (AMF-GAM) is shown in Fig.~\ref{fig5} as an example. Firstly, we implement spatial-temporal joint graph convolution for the input ${X^l} \in {{\mathbb{R}}^{N \times T \times {C_{\text{i}}}}}$ with $B$ STConv kernels to extract multiple spatial-temporal features. Specifically the inception is implemented with padding, so the output of each branch $b$ can be ${F_b} \in {{\mathbb{R}}^{N \times T \times K \times {C_o}}}$. Then extracted features of $B$ branches are concatenated to form the output ${{B}_{out}}\in {{\mathbb{R}}^{N\times T\times K\times ({C_{o}}\times B)}}$, where we set the number of output channels ${C_{o}}$ of each branch to be the same.

Here we introduce the graph node attention, which has being widely used \cite{2017Attention}. To solve the heterogeneity, we use the $Q={{S}_{e}}{{W}_{q}}$ as graph node query, where $Q \in {{\mathbb{R}}^{N \times {C_o}}}$, ${{W}_{q}}\in {{\mathbb{R}}^{d\times C_o}}$. For each branch output ${F_b}$, we apply global pooling \cite{2019Selective} $Ke{y_b} = \sum\limits_{i = 1}^T {\sum\limits_{j = 1}^K {{F_b}} } $, where $Ke{y_b} \in {{\mathbb{R}}^{N \times C_o}}$ as the corresponding key of branch $b$, then $Key \in {{\mathbb{R}}^{N \times (C_o \times B)}}$ for $B$ branches (one can also use ${W_k} \in {{\mathbb{R}}^{C_o \times C_o}}$ to do transform, here we omit it for simplicity). Then we compute the attention score by $A = Q{Key^T}$, where $A \in {{\mathbb{R}}^{N \times B}}$. Take a graph node ${{v}_{i}}$ as an example, ${A_{b,{v_i}}} = \frac{{{Q_{{v_i}}} \cdot {Key_{b,{v_i}}}}}{{\sqrt C_o }}$, where ${Q_{{v_i}}} \in {{\mathbb{R}}^{1 \times C_o}}$, ${Key_{b,{v_i}}} \in {{\mathbb{R}}^{1 \times C_o}}$, denotes the attention of ${{v}_{i}}$ with each branch extracted spatial-temporal features. Then we concatenate results of each branch $F_b$ adjusted by attention score $As$ to obtain the final AMF-GAM output $Ao$. The calculation can be formulated as follows:

\begin{equation}
	A{s_{{v_i},b}} = \frac{{\exp ({A_{b,{v_i}}})}}{{\sum\limits_{r = 1}^B {\exp ({A_{r,{v_i}}})} }}
\end{equation}

\begin{equation}
	A{o_{{v_i}}} = ||_{b = 1}^B\left\{ {{A_{b,{v_i}}} \cdot {F_{b,{v_i}}}} \right\}
\end{equation}

\noindent where ${A_{b,{v_i}}} \in {\mathbb{R}}$, ${F_{b,{v_i}}} \in {{\mathbb{R}}^{T \times K \times {C_o}}}$, $A{o_{{v_i}}} \in {{\mathbb{R}}^{T \times K \times {C_o} \times B}}$ represent the output with node ${{v}_{i}}$ of the AMF-GAM. Therefore, the final output of the AMF-GAM is $Ao \in {{\mathbb{R}}^{N \times T \times K \times {C_o} \times B}}$.

\textbf{AMF-STConv block output Layer} 

Then, we use a two-layer fully connected neural network to generate the output of AMF-STConv block. We first reshape ${Ao}$ into ${Ao}\in {{\mathbb{R}}^{N\times T\times K\times (C_o\times B)}}$. Next, the learnable weight matrix ${{W}_{s}}\in {{\mathbb{R}}^{K\times \left( C_o\times B \right)\times (C_o\times B)}}$ is used to convert ${Ao}$ to ${Ao}\in {{\mathbb{R}}^{N\times T\times (C_o\times B)}}$. We also implement SE-net \cite{SENET} to model the channel attention. Finally, the output is converted to ${X^{l + 1}}\in {{\mathbb{R}}^{N\times T\times C_o}}$ using the second full connection layer with ${{W}_{o}}\in {{\mathbb{R}}^{\left( C_o\times B \right)\times C_o}}$. And the process can be formulated as ${X^{l + 1}}={Ao}{{W}_{s}}{{W}_{o}}$. In addition, we implement residual networks \cite{Residual} and layer normalization to improve the network performance.

\subsection{Fusion Output Module}

After pre-trained in the AMF-STConv block, we propose two decoder modules to fuse features: Time-Step decoder and Fully-Connected decoder, as shown in Fig.~\ref{fig3}. In Time-Step decoder, each forecasted time step is concatenated with the inputs along with previous observations to forecast next step, which called iterative forecasting. However, in multi-step traffic forecasting, iterative forecasting is prone to error accumulation, which leads to the gradual deterioration of the long-term forecasting accuracy. Fully-Connected decoder uses two fully-connected layers to forecast multi-step results at the same time. But the short-term forecasting performance of Fully-Connected decoder is less good than Time-Step decoder. To benefit from both, we fuse the outputs of Time-Step decoder and Fully-Connected decoder to obtain better forecasting performance. At last, we adopt the fusion of two decoders as:

\begin{equation}
    \hat Y = {W_f} \odot {\hat Y_{{fc}\_out}} + (1 - {W_f}){\hat Y_{ts\_out}}
\end{equation}

\noindent where ${{\hat{Y}}_{ts\_out}}$, ${{\hat{Y}}_{fc\_out}} \in {{\mathbb{R}}^{N\times M\times C}}$ represent the forecasting of Time-Step decoder and Fully-Connected decoder respectively. ${W_f}$ is learning parameters, $\hat{Y}$ is the final multi-step prediction results.

In this paper, we use the mean square error (MSE) as the loss function and minimize it through backpropagation.

\begin{equation}
    {\text{L(}}\theta {\text{) = }}\frac{1}{M}\sum\limits_{i = t + 1}^{t + M} {({Y_i} - } {\hat Y_i}{)^2}
\end{equation}

\noindent where $\theta $ represents all learnable parameters of our model, ${{\hat{Y}}_{i}}$ denotes the model’s forecasting results of all nodes at time step $i$, ${{Y}_{i}}$ is the ground truth.

\section{Experiment}

\subsection{Datasets}

We evaluate AMF-STGCN on two mobile traffic datasets (Milan, Jiangsu) and two road traffic datasets (PEMS04, PEMS08). Milan dataset comes from the mobile traffic volume records of Milan provided by Telecom Italia \cite{2015Milan}. Jiangsu dataset is mobile traffic volume sampled from 1051 cells in Jiangsu province. The road traffic datasets come from the Caltrans Performance Measurement System (PeMS) \cite{2001Freeway}. We summarize the statistics of the datasets in Table~\ref{tab1}.

\begin{table}[htbp]
  \centering
  \caption{Dataset description}
  \begin{tabular}{p{0.8cm}<{\centering}p{0.8cm}<{\centering}p{0.5cm}<{\centering}p{1.7cm}<{\centering}p{0.8cm}<{\centering}p{0.6cm}<{\centering}p{0.7cm}<{\centering}}
    \toprule
    \textbf{Datasets} & \textbf{Samples} & \textbf{Nodes} & \textbf{Timespan} &  \textbf{Timeslot} & \textbf{Input Length} & \textbf{Output Length} \\
    \midrule
    Milan & 4320  & 900 & Nov, 2013 & 10min & 6     & 6 \\
    \midrule
    Jiangsu & 8640  & 1051 & Jan-Mar, 2021 & 15min & 12     & 12 \\
    \midrule
    PEMS04 & 16992 & 307 & Jan-Feb, 2018   & 5min  & 12    & 12 \\
    \midrule
    PEMS08 & 17856 & 170 & Jul-Aug, 2016  & 5min  & 12    & 12 \\
    \bottomrule
    \end{tabular}%
  \label{tab1}%
\end{table}%

The ratios of train set, validation set and test set on four datasets are 2:0:1, 2:1:1, 6:2:2 and 6:2:2. All measurements are normalized to [0,1]. In the experiment, historical time window and the forecasting window on Milan dataset are set to 6, other datasets are set to 12. The hyperparameters of AMF-STGCN are determined by the performance on the validation set. The size of graph convolution kernels are set to 3$\times$1, 1$\times$3, 5$\times$2, 3$\times$2, 2$\times$3, which are used in spatial-temporal inception branches respectively. We evaluate the performance of different methods using Mean Absolute Error (MAE) and Root Mean Square Error (RMSE). Here, all results are denormalized before evaluation. 

\subsection{Experimental Settings}

For Jiangsu, PEMS04 and PEMS08, adjacency matrix and parameters are same as in STGCN \cite{yu2017spatio}. For Milan dataset, Spearman correlation coefficient was used to define the adjacency matrix, and threshold is set to 0.92.

We compare AMF-STGCN with other widely used baseline models, including HA (Historical Average method), ARIMA (Auto-Regressive Integrated Moving Average), LSTM \cite{1997Long}, STGCN \cite{yu2017spatio}, ASTGCN \cite{guo2019attention}, AGCRN \cite{AGCRN} and STSGCN \cite{song2020spatial}.

\subsection{Comparison and Result Analysis}

\begin{table*}[tbp]
  \centering
  \caption{Performance comparison of different approaches}
    \begin{tabular}{p{7.00em}cccccccc}
    \toprule
    \textbf{Dataset} & \multicolumn{2}{c}{\textbf{Milan}} & \multicolumn{2}{c}{\textbf{Jiangsu}} & \multicolumn{2}{c}{\textbf{PEMS04}} & \multicolumn{2}{c}{\textbf{PEMS08}} \\
    \midrule
    \textbf{Model} & \multicolumn{1}{c}{\textbf{MAE}} & \multicolumn{1}{c}{\textbf{RMSE}} & \multicolumn{1}{c}{\textbf{MAE}} & \multicolumn{1}{c}{\textbf{RMSE}} & \multicolumn{1}{c}{\textbf{MAE}} & \multicolumn{1}{c}{\textbf{RMSE}} & \multicolumn{1}{c}{\textbf{MAE}} & \multicolumn{1}{c}{\textbf{RMSE}} \\
    \midrule
    \textbf{HA} & 61.28 & 120.73 & 184.94 & 358.00 & 61.80 & 87.27 & 31.94 & 46.34 \\
    \midrule
    \textbf{ARIMA} & 54.61 & 78.94 & 174.13 & 445.47 & 32.11 & 68.13 & 24.04 & 43.30 \\
    \midrule
    \textbf{LSTM} & 42.93 & 80.29 & 198.37 & 300.95 & 29.04 & 45.63 & 25.93 & 38.64 \\
    \midrule
    \textbf{STGCN} & \underline{34.77} & \underline{63.28} & 175.81 & 306.24 & 22.15 & 34.68 & 18.49 & 28.18 \\
    \midrule
    \textbf{ASTGCN} & 39.42 & 72.04 & 164.28 & 300.20 & 22.99 & 35.42 & 18.89 & 28.64 \\
    \midrule
    \textbf{AGCRN} & 42.27 & 135.18 & \underline{131.83} & \underline{279.51} & \underline{20.11} & \underline{32.88} & \underline{16.19} & \underline{25.56} \\
    \midrule
    \textbf{STSGCN} & \textbackslash & \textbackslash & \textbackslash & \textbackslash & 21.29 & 33.81 & 17.18 & 26.85 \\
    \midrule
    \textbf{AMF-STGCN} & \textbf{29.84} & \textbf{57.73} & \textbf{129.28} & \textbf{247.41} & \textbf{19.45} & \textbf{31.57} & \textbf{15.53} & \textbf{24.60} \\
    \midrule
    \textbf{Improvements} & +14.18\% & +8.77\% & +1.94\% & +11.48\% & +3.28\% & +3.98\% & +4.08\% & +3.76\% \\
    \bottomrule
    \end{tabular}%
  \label{tab2}%
\end{table*}%

Table~\ref{tab2} shows the overall performance on four real-world datasets. Due to the huge computation cost, it is hard to measure the performance of STSGCN on mobile traffic datasets with more nodes. It is found that AMF-STGCN achieves state-of-the-art results on all datasets. The improvements of MAE on four datasets are between 1.94\% and 14.18\%, and the improvements of RMSE range from 3.76\% to 11.48\%. This also shows that AMF-STGCN is more robust in spatial-temporal forecasting tasks.

The performance improvements of AMF-STGCN on mobile traffic datasets are much greater than road traffic datasets. The improvements of MAE and RMSE on the mobile traffic datasets are 14.18\% and 11.48\%, while 4.08\% and 3.98\% on road traffic datasets. Based on previous data analysis results, we found that mobile traffic datasets have more obvious heterogeneity than road traffic datasets. Our model has more improvements in mobile traffic datasets, proving that it can extract spatial-temporal heterogeneity.

The methods based on GCN (STGCN, ASTGCN, AGCRN, STSGCN, AMF-STGCN) are more effective than time series forecasting methods (HA, ARIMA, LSTM), which illustrates the limitations of time series forecasting methods in modeling spatial-temporal data. Since STGCN and ASTGCN model the spatial-temporal dependencies separately and do not take heterogeneity into account, their effects are worse than AGCRN, STSGCN and AMF-STGCN in most datasets. Because AGCRN is limited by the dimension of node embedding, its generalization ability is weak. Besides, STSGCN only considers local spatial-temporal heterogeneity. Both of those methods are difficult to achieve optimal results. However, our AMF-STGCN can model spatial-temporal dependencies and nodes' heterogeneity at the same time, so it has advantages in modeling complex scenarios of traffic forecasting.

\begin{figure}[tbph]
\begin{center}
\includegraphics[width=8.5cm]{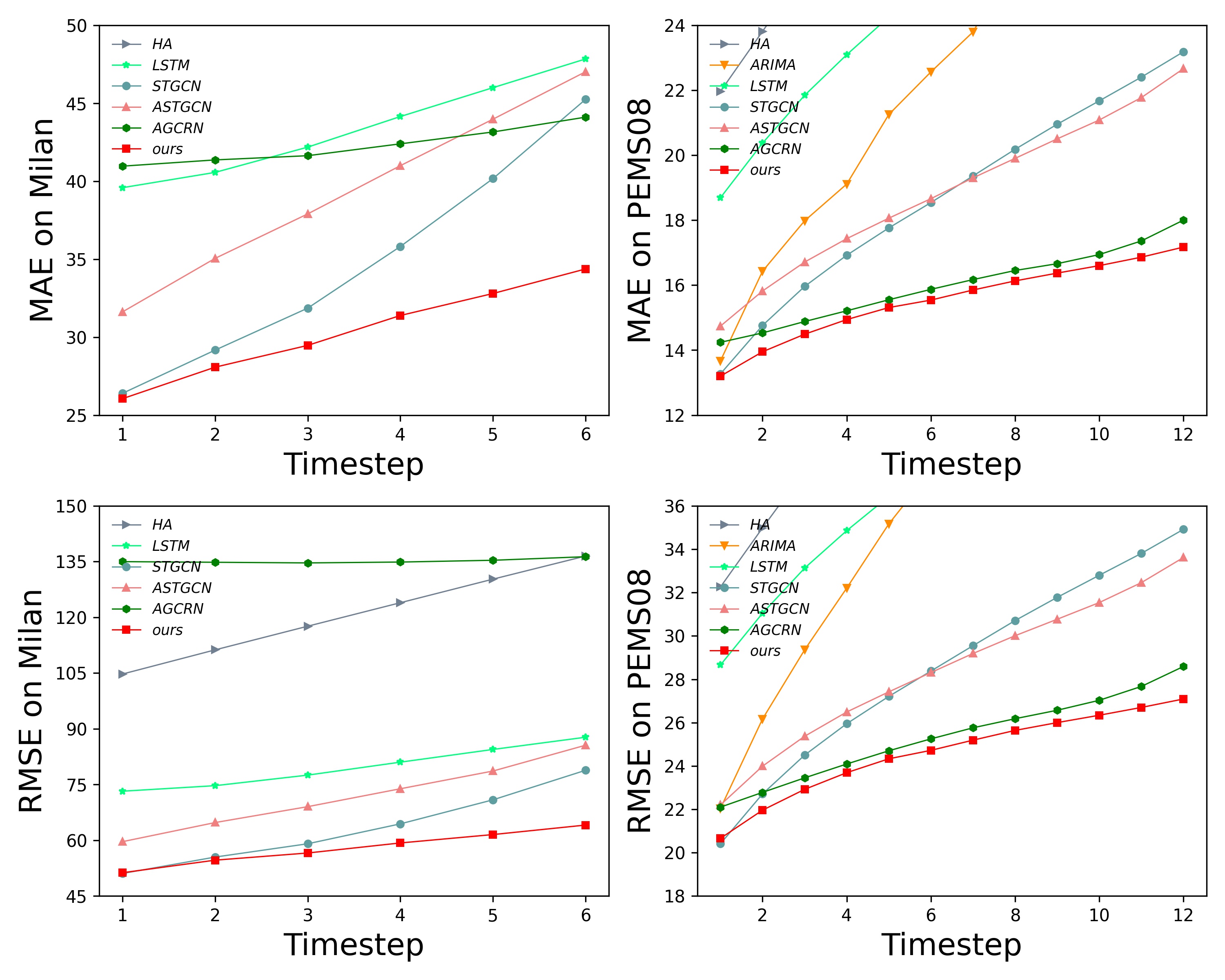}
\end{center}
\caption{Forecasting performance comparison at each time step.}
\label{fig6}
\end{figure}

Fig.~\ref{fig6} shows changes of different metrics on the Milan and PEMS08 datasets with increasing prediction length. The metrics value of HA is too large to be displayed completely. As we can see from the Fig.~\ref{fig6}, the error increases over time, indicating that forecasting task becomes more difficult. The performance of GCN-based methods is stable, which shows the effectiveness of capturing spatial information. Although our model has no outstanding performance in short-term forecasting tasks, it shows great advantages in medium and long-term forecasting tasks. This benefits from Fully-Connected decoder which alleviates the error accumulation of multi-step forecasting.

\subsection{Ablation Study}

In this section, we further carry out ablation tests on Jiangsu and PEMS08 datasets to study the effectiveness of each key component used in AMF-STGCN, including: (1) AMF-STConv block, (2) AMF-Attention, (3) mask. Except for the different control variables, the settings of each experiment were the same.

Table~\ref{tab3} shows performance of AMF-STGCN against its ablations. We can observe that: (1) Replacing AMF-STConv block with spatial-temporal convolutional block of STGCN (w/o AMF-STConv block), the model performs worst compared with other ablation models. The result proves the superior ability of AMF-STConv block to capture heterogeneity of graph nodes and model spatial-temporal dependencies simultaneously. (2) Without attention in AMF-STConv block (w/o AMF-Att), the performance is second worst. We believe that model cannot extract heterogeneity. Take Jiangsu dataset as an example, the attention scores of Node0 and Node1 are [0.0015, 0.94, 0.003, 0.057] and [0.015, 0.31, 0.029, 0.65] respectively. It means Node0 is most affected by the second STConv kernel, whose effective affected range is within 5 hops in spatial dimension, and 2 time steps in temporal dimension. While Node1 is most affected by the fourth STConv kernel, followed by the second
\begin{table}[tbph]
  \centering
  \caption{Ablation Results of AMF-STGCN}
    \begin{tabular}{l|cccc}
    \toprule
    \textbf{Dataset} & \multicolumn{2}{c}{\textbf{Jiangsu}} & \multicolumn{2}{c}{\textbf{PEMS08}} \\
    \midrule
    \textbf{Model} & \textbf{MAE} & \textbf{RMSE} & \textbf{MAE} & \textbf{RMSE} \\
    \midrule
    AMF-STGCN  & 129.28 & 247.42 & 9.71 & 15.53 \\
    \midrule
    w/o AMF-STConv block & 141.25 & 263.73 & 10.53 & 17.01 \\
    \midrule
    w/o AMF-Att  & 134.64 & 252.95 & 9.94 & 16.18\\
    \midrule
    w/o mask & 129.93 & 248.07 & 9.73 & 15.56 \\
    \bottomrule
    \end{tabular}%
  \label{tab3}%
\end{table}%

\noindent kernel. (3) With the addition of the mask mechanism, the MAE is reduced by 0.56 and 0.2, and the RMSE is reduced by 0.55 and 0.03, respectively. This shows that the mask mechanism can further improve the robustness of the model.

\subsection{Computational Complexity}

We compare computational complexity between AMF-STGCN, STGCN, ASTGCN, AGCRN and STSGCN on the PEMS04 dataset. The results are shown in Table~\ref{tab4}. The experiments are conducted on Tesla V100 server. STGCN uses a fully convolutional structure, so its training speed is fastest. ASTGCN introduces temporal and spatial attention on the basis of STGCN to capture dynamic spatial-temporal dependencies, so the amount of parameters and training time are greatly increased. Compared with AMF-STGCN, the above two methods have limitations in terms of interpretability and generalization ability. AGCRN introduces a large number of parameters, as it uses learnable node embedding to represent node status. In addition, AGCRN uses recurrent structures to model temporal dependence, which is more time-consuming. STSGCN constructs STSGC modules based on the complex localized spatial-temporal graphs to model localized spatial-temporal correlations. The parameters of STSGCM layers are not shared, resulting in a linear increase in the number of parameters. Compared with the aforementioned methods, AMF-STGCN is more suitable for applications.

At present, the proposed method has been deployed in the network of a certain province in China. By introducing the spatial-temporal forecasting model into operation and maintenance process, operators can identify cells with high volume of business in advance and plan the maintenance priority of these cells.

\begin{table}[bthp]
  \centering
  \caption{The computational Complexity on the PeMS04 dataset}
    \begin{tabular}{l|cc}
    \toprule
    Models & \multicolumn{1}{l}{Parameters} & \multicolumn{1}{l}{Training time(s/epoch)} \\
    \midrule
    AMF-STGCN & 677591 & 45.88 \\
    \midrule
    STGCN & 211596 & 10.81 \\
    \midrule
    ASTGCN & 450031 & 26.74 \\
    \midrule
    AGCRN & 748810 & 72.38 \\
    \midrule
    STSGCN & 2872686 & 109.93 \\
    \bottomrule
    \end{tabular}%
  \label{tab4}%
\end{table}%

\section{Conclusion}


In this paper, we elaborate on a new GCN-based model, AMF-STGCN. We evaluate our method on four real-world datasets from two different domains, and obtain consistent better results comparing to various advance models in the literature of  prediction. In the future, it is valuable to model dynamics and abrupt changes of mobile traffic data, as well as solve the massive node scenarios problem.






\end{document}